\pdfoutput=1

\def\year{2020}\relax
\documentclass[letterpaper]{article} 
\usepackage{aaai20}  
\usepackage{times}  
\usepackage{helvet} 
\usepackage{courier}  
\usepackage[hyphens]{url}  
\usepackage{graphicx} 
\usepackage{graphicx}  
\usepackage{xcolor}
\usepackage{amsmath}
\usepackage{booktabs}
\usepackage{bm}
\usepackage{mathtools}
\usepackage[multiple]{footmisc}
\usepackage{textcomp}
\usepackage{subcaption}

\newenvironment{flushitemize}{%
\begin{list}{$\bullet$}
   {\setlength{\leftmargin}{15pt}}%
    \setlength{\labelwidth}{20pt}
    \setlength{\itemindent}{0pt}
    \setlength{\labelsep}{0.5em}
 \setlength{\itemsep}{1pt}
 \setlength{\parskip}{0pt} 
 \setlength{\parsep}{0pt}}
 {\end{list}}

\title{Path Ranking with Attention to Type Hierarchies}
\author{Weiyu Liu, Angel Daruna, Zsolt Kira, Sonia Chernova\\
Institute for Robotics and Intelligent Machines\\
Georgia Institute of Technology, Atlanta, Georgia\\
\{wliu88,adaruna3,zkira\}@gatech.edu, chernova@cc.gatech.edu
}

\begin{document}

\maketitle

\begin{abstract}
The objective of the knowledge base completion problem is to infer missing information from existing facts in a knowledge base. Prior work has demonstrated the effectiveness of path-ranking based methods, which solve the problem by discovering observable patterns in knowledge graphs, consisting of nodes representing entities and edges representing relations. However, these patterns either lack accuracy because they rely solely on relations or cannot easily generalize due to the direct use of specific entity information. We introduce Attentive Path Ranking, a novel path pattern representation that leverages type hierarchies of entities to both avoid ambiguity and maintain generalization. Then, we present an end-to-end trained attention-based RNN model to discover the new path patterns from data. Experiments conducted on benchmark knowledge base completion datasets WN18RR and FB15k-237 demonstrate that the proposed model outperforms existing methods on the fact prediction task by statistically significant margins of $26\%$ and $10\%$, respectively. Furthermore, quantitative and qualitative analyses show that the path patterns balance between generalization and discrimination.
\end{abstract}

\section{Introduction}

Knowledge bases (KBs), such as WordNet \cite{wordnet} and Freebase \cite{freebase}, have been used to provide background knowledge for tasks such as recommendation \cite{wang2018dkn}, and visual question answering \cite{aditya2018explicit}. Such KBs typically contain facts stored in the form of $(source\:entity,\, relation,\, target \:entity)$ triples, such as $(Fork, in, Kitchen)$. Combined, the dataset of triples is often represented as a graph, consisting of nodes representing entities and edges representing relations. Many KBs also contain type information for entities, which can be represented as type hierarchies by ordering each entity's types based on levels of abstraction. 

Despite containing millions of facts, existing KBs still have a large amount of missing information \cite{min2013distant}. As a result, robustly reasoning about missing information is important not only for improving the quality of KBs, but also for providing more reliable information for applications relying on the contained data. The objective of the \textit{knowledge base completion problem} is to infer missing information from existing facts in KBs. More specifically, \textit{fact prediction} is the problem of predicting whether a missing triple is true. 

\begin{figure}[t!]
  \includegraphics[width=1.0\linewidth]{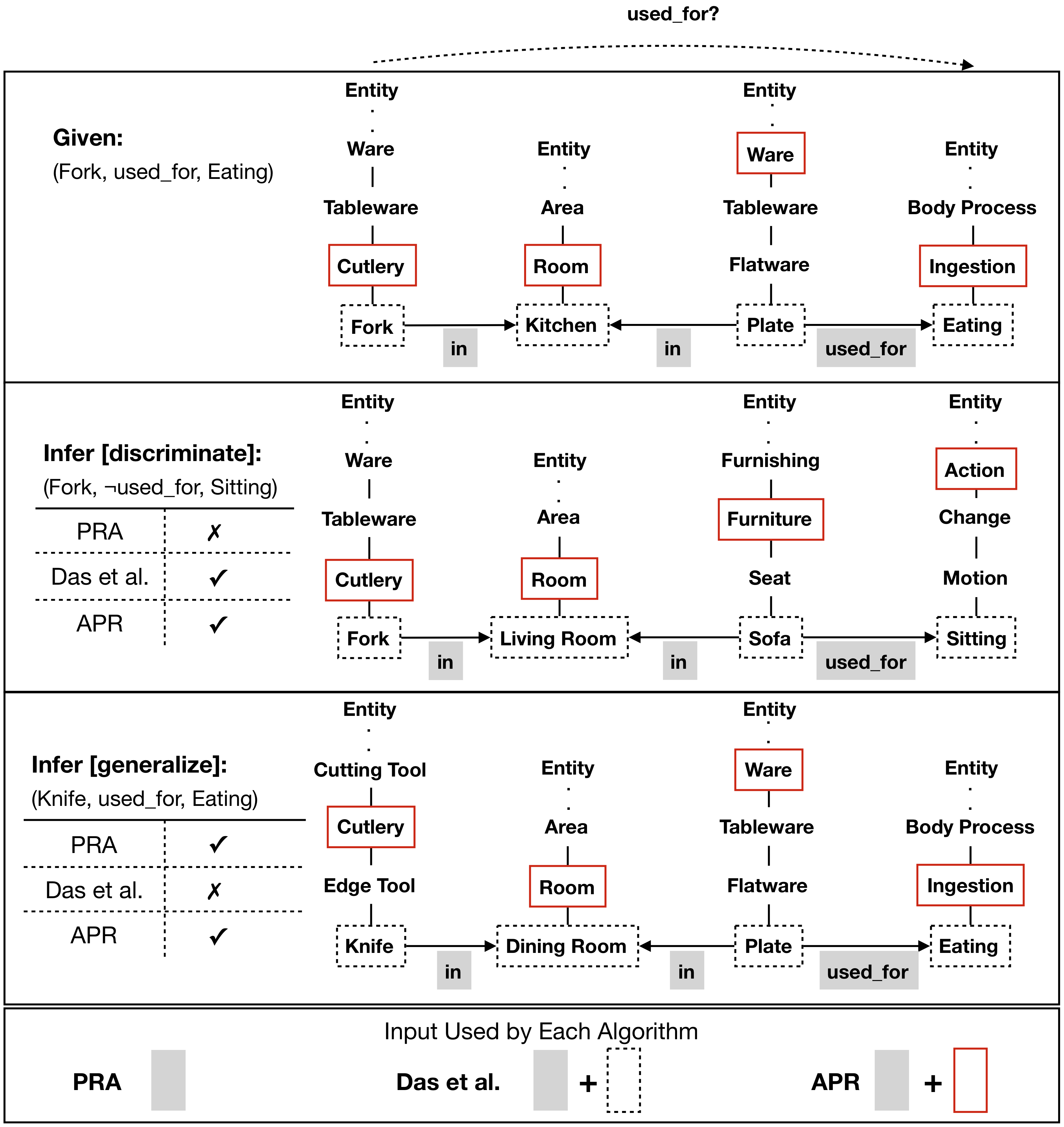}
  \centering
  \caption{Paths and type hierarchies of a known triple and two missing triples. 
  Our proposed method, Attentive Path Ranking, can correctly predict both missing triples by using attention to select types at appropriate levels of abstraction.}
  \label{fig:main_figure}
\end{figure}

Prior work on fact prediction has demonstrated the effectiveness of path-ranking based methods \cite{pra,sfe,cvsm,chains}, which solve the knowledge base completion problem by discovering observable patterns in knowledge graphs. A foundational approach in this area is the Path Ranking Algorithm (PRA) \cite{pra}, which extracts patterns based on sequences of relations. However, relying on relation-only patterns often leads to overgeneralization, as shown in the example in Figure \ref{fig:main_figure}. In this example, the top triple ($\langle Fork, used\_for, Eating \rangle$) and its associated information is assumed to be known, and the second ($\langle Fork, \lnot used\_for, Sitting \rangle$) and third ($\langle Knife, used\_for, Eating \rangle$) triples must be inferred from available information. PRA is limited to using only relation patterns, and as such uses $\langle in,-in,used\_for\rangle$\footnote{the negative sign indicates reversed direction for the relation}, observed in the path of the existing triple, to make its predictions. As the example shows, this representation enables PRA to correctly generalize to similar scenarios, such as the third triple, but the lack of context leads to over-generalization and failure to discriminate the second triple.

To improve the discriminativeness of PRA, \cite{chains} extend the above approach to incorporate entity information in addition to relations within the learned paths. Paths are expanded to include entities directly, or by aggregating all types for each entity. As an example, the path pattern of the known triple in Figure \ref{fig:main_figure} becomes $\langle Fork, in, Kitchen, -in, Plate, used\_for, Eating \rangle$. This new approach successfully discriminates the second triple, however, this comes at the cost of losing generalizability, resulting in misclassification of the third triple.

In our work, we introduce a novel path-ranking approach, \textbf{Attentive Path Ranking (APR)} \footnote{The code and data are available at: \url{https://github.com/wliu88/AttentivePathRanking.git}}, that seeks to achieve discrimination and generalization simultaneously.  Our work is motivated by the distributional informativeness hypothesis \cite{santus2014hypernym}, according to which the generality of a term can be inferred from the informativeness of its most typical linguistic contexts.
Similar to \cite{chains}, we utilize a path pattern consisting of both entity types and relations.  However, for each entity
we use an \textit{attention mechanism} \cite{ntm} to select a single type representation for that entity at the appropriate level of abstraction, thereby achieving both generalization and discrimination. As shown in Figure~\ref{fig:main_figure}, the APR path pattern of the known triple becomes $\langle Cutlery, in, Room, -in, Ware, used\_for, Ingestion \rangle$, which successfully discriminates from the second triple, and correctly generalizes to the third.  


Our work makes the following contributions:  
\begin{flushitemize}
    \item We introduce an end-to-end trained attention-based RNN model for entity type selection. Our approach builds on prior work on attention mechanisms, while contributing a novel approach for learning the appropriate levels of abstraction of entities from data. 
    \item Based on the above, we introduce Attentive Path Ranking, a novel path pattern representation that leverages type hierarchies of entities to both avoid ambiguity and maintain generalization. 
    \item Applicable to the above, and other path-ranking based methods more broadly, we introduce a novel pooling method based on attention to jointly reason about contextually important information from multiple paths.
\end{flushitemize}
We quantitatively validate our approach against five prior methods on two benchmark datasets: WN18RR \cite{wn18rr} and FB15k-237 \cite{fb15k-237}.  Our results show statistically significant improvement of $26\%$ on WN18RR and $10\%$ on FB15k-237 over state-of-the-art path-ranking techniques.  Additionally, we demonstrate that our attention method is more effective than fixed levels of type abstraction, and that attention-based pooling improves performance over other standard pooling techniques.  Finally, we show that APR significantly outperforms knowledge graph embedding methods on this task.

\section{Related Work}

In this section, we present a summary of prior work. 

\subsection{Knowledge Based Reasoning}
Multiple approaches to knowledge based reasoning have been proposed. Knowledge graph embedding (KGE) methods map relations and entities to vector representations in continuous spaces \cite{nickel}. Methods based on inductive logic programming discover general rules from examples \cite{quinlan1993foil}. Statistical relational learning (SRL) methods combine logics and graphical models to probabilistically reason about entities and relations \cite{getoor2007srl}. Reinforcement learning based models treat link prediction (predicting target entities given a source entity and a relation) as Markov decision processes \cite{deeppath,das2017go}. Path-ranking based models use supervised learning to discover generalizable path patterns from graphs \cite{pra,sfe,cvsm,chains}.

In the context of knowledge base completion, we focus on path-ranking based models. The Path Ranking Algorithm (PRA) \cite{pra} is the first proposed use of patterns based on sequences of relations.  By using patterns as features of entity pairs, the fact prediction problem is solved as a classification problem.  However, the algorithm is computationally intensive because it uses random walks to discover patterns and calculate feature weights.  Subgraph Feature Extraction (SFE) \cite{sfe} reduces the computational complexity by treating patterns as binary features, as feature weights provide no discernible benefit to the performance, and using more efficient bi-directional breadth-first search to exhaustively search for sequences of relations and additional patterns in graphs.  To generalize semantically similar patterns, \cite{cvsm} use recurrent neural networks (RNNs) to create vector representations of patterns, which are then used as multidimensional features for prediction. \cite{chains} improve the accuracy of the RNN method by making use of the additional information in entities and entity types. We also use entity types but we focus on using types at different levels of abstraction for different entities.


\subsection{Representing Hierarchical Structures}

\begin{figure*}[t]
  \includegraphics[width=\linewidth]{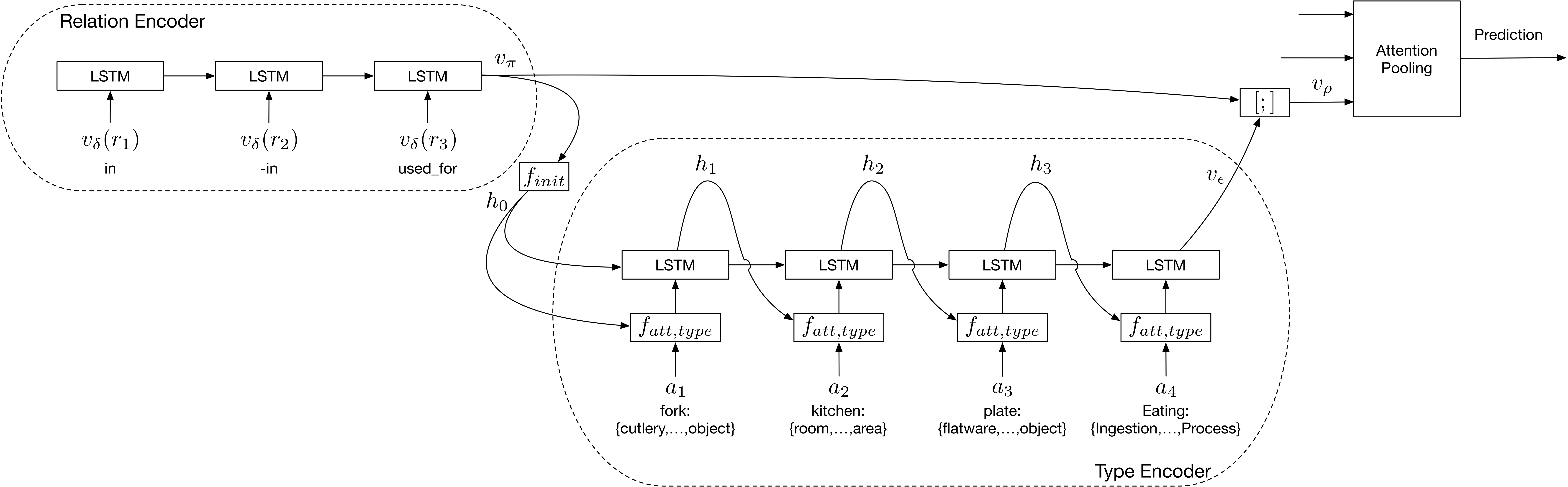}
  \centering
  \caption{Relation encoder and entity type encoder with attention to type hierarchies.}
  \label{fig:Network Architecture}
\end{figure*}

Learning representations of hierarchical structures in natural data such as text and images has been shown to be effective for tasks such as hypernym classification and textual entailment. \cite{vendrov2016oe} order text and images by mapping them to a non-negative space in which entities that are closer to the origin are more general than entities that are further away. \cite{athiwaratkun2018hierarchical} use density order embeddings where more specific entities have smaller, concentrated probabilistic distributions and are encapsulated in broader distributions of general entities. In this work, we do not explicitly learn the representation of hierarchical types. Instead, we leverage the fact that types in type hierarchies have different levels of abstraction to create path patterns that balance generalization and discrimination. 

\subsection{Attention}

Attention was first introduced in \cite{ntm} for machine translation, where it was used to enable the encoder-decoder model to condition generation of translated words to different parts of the original sentence. Later, cross-modality attention was shown to be effective at image captioning \cite{xu2015show} and speech recognition \cite{chan2015listen}. Our approach uses attention to focus on contextually important information from multiple paths, much like the above methods. More importantly, we use attention in a novel way to efficiently discover the correct levels of abstraction for entities from a large search space. 

\section{Problem Definition}

A KB is formally defined as a set of triples, also called relation instances, $\mathcal{X} = \{ (e_i, r_j, e_k)|e_i, e_k\in \mathcal{E} \land  r_j \in \mathcal{R} \}$, where $\mathcal{E}$ denotes the entity set, and $\mathcal{R}$ denotes the relation set. The KB can be represented as a multi-relational graph $\mathcal{G}$, where nodes are entities and edges are relations. A directed edge from $e_i$ to $e_k$ with label $r_j$ exists for each triple $(e_i, r_j, e_k)$ in $\mathcal{X}$.
A path between $e_i$ and $e_k$ in $\mathcal{G}$ is denoted by $p = \langle e_1,r_1,...,r_M,e_{M+1}\rangle$, where $e_1 = e_i$ and $e_{M+1} = e_k$. The length of a path is defined as the number of relations in the path, $M$ in this case. For all pairs of entities $e_i$ and $e_k$ in the graph $\mathcal{G}$, we can discover a set of $N$ paths up to a fixed length, $\mathcal{P}_{ik} = \{ p_1,...,p_N \}$. 

Our objective is, given an incomplete KB and the path set $\mathcal{P}_{ik}$ extracted from the KB, to predict whether the missing triple $(e_i, r_j, e_k)$ is true, or equivalently whether the entity pair $e_i$ and $e_k$ can be linked by $r_j$.

\section{Attentive Path Ranking}

In this section, we present our proposed Attentive Path Ranking model, which takes as input the set of paths between entities $e_i$ and $e_k$, $\mathcal{P}_{ik}$, and outputs $P(r_j|e_i, e_k)$, the probability that $r_j$ connects $e_i$ and $e_k$. Our model, shown in Figure \ref{fig:Network Architecture}, consists of three components: a relation encoder, an entity type encoder, and an attention-based pooling method. Given a path $p = \langle e_1,r_1,...,r_M,e_{M+1}\rangle$ in $\mathcal{P}_{ik}$, the \textbf{relation encoder} encodes the sequence of relations $\langle r_1,...,r_M\rangle$ (e.g., $\langle in, -in, used\_for \rangle$ in Figure \ref{fig:Network Architecture}). The \textbf{entity type encoder} both selects a type $l_t$ for each entity $e_t$ with a certain level of abstraction, and encodes the selected types $\langle l_1,...,l_{M+1} \rangle$ (e.g., $\langle Cutlery,\, Room,\, Ware,\, Ingestion \rangle$). We then combine the relation and entity type encodings to form the path pattern $\hat{p} = \langle l_1,r_1,...,r_M,l_{M+1}\rangle$. The above process is repeated for each path in $\mathcal{P}_{ik}$. \textbf{Attention-based pooling} is then used to combine all path patterns to predict the truth value of $(e_i, r_j, e_k)$. 
In the following sections we present details of the three core model components.

\subsection{Relation Encoder}

The relation encoder uses a LSTM \cite{lstm} to sequentially encode vector embeddings of relations $\bm{v}_\delta(r_t)$ for all relations in the path $p$. Here, the trainable vector embeddings help generalize semantically similar relations by representing them as similar vectors. The last state of the LSTM is used as the vector representation of the sequence of relations, denoted by $\bm{v}_\pi(\hat{p})$. We use a LSTM instead of a simple RNN for its ability to model long-term dependencies, which aids in modeling longer paths.

\subsection{Entity Type Encoder}

The entity type encoder consists of attention modules applied to entity types and a second LSTM. 
Together, these models are responsible for selecting a type $l_t$ for each entity $e_t$ from its type hierarchy $\mathcal{L}_t = \{ l_{t,1},...,l_{t,C}\}$, where the lowest level $l_{t,1}$ represents the most specific type and the highest level $l_{t,C}$ represents the most abstract type in a hierarchy of height $C$.

As shown by the distributional informativeness hypothesis \cite{santus2014hypernym}, selecting $l_t$ from more specific levels of the type hierarchy increases the discriminativeness of the path pattern $\hat{p}$ while selecting from more abstract levels makes the path pattern easier to generalize. Choosing $l_t$ at the appropriate level helps create a path pattern that is both discriminative and generalizable, leading to greater prediction accuracy. However, the substantial number of combinations when considering possible types for entities in all path patterns makes exhaustively searching across all $l_t$'s impossible.



To select $l_t$, we use the deterministic ``soft" attention introduced in \cite{ntm} to create an approximated vector representation of $l_t$ from the set of type vectors $\{\bm{v}_\tau(l_{t,1}),...,\bm{v}_\tau(l_{t,C})\}$, where $\bm{v}_\tau(l_{t,i})$ can be obtained by learning vector embeddings of entity types. We name this approximated vector representation of $l_t$ as the type context vector $\hat{\mathbf{a}}_t$. For each type vector $\bm{v}_\tau(l_{t,i})$ of entity $e_t$, a weight $\alpha_{t,i}$ is computed by a feed-forward network $f_{att, type}$ conditioned on an evolving \textbf{context} $\hat{\mathbf{c}}_t$. This weight can be interpreted as the probability that $l_{t,i}$ is the right level of abstraction or the relative importance for level $i$ to combine $\bm{v}_\tau(l_{t,i})$'s together. Formally, $\hat{\mathbf{a}}_t$ can be calculated as:%
\begin{align}
	e_{t,i} =& f_{att, type}(\bm{v}_\tau(l_{t,i}), \hat{\mathbf{c}}_t) \\
	\alpha_{t,i} =& \dfrac{exp(e_{t,i})}{\sum_{k=1}^{C} exp(e_{t,k})}\\
	\hat{\mathbf{a}}_t =& \sum_{i=1}^{C} \alpha_{t,i} \bm{v}_\tau(l_{t,i})
\end{align} %

We model the context of the current step as the previous hidden state of the LSTM, i.e., $\hat{\mathbf{c}}_t = \bm{h}_{t-1}$. 
We also use $\bm{v}_\pi(\hat{p})$, the last state of the relation encoder, to compute the initial memory state and hidden state of the LSTM:%
\begin{align}
	\bm{c}_0 =& f_{init, c}(\bm{v}_\pi(\hat{p})) \\
	\bm{h}_0 =& f_{init, h}(\bm{v}_\pi(\hat{p}))
\end{align} %
where $f_{init, c}$ and $f_{init, h}$ are two separate feed-forward networks. Illustrated in Figure \ref{fig:Network Architecture}, as the LSTM stores information from the relation encoder and previously approximated entity types, both the sequence of relations $\langle in, -in, used\_for \rangle$ and the type context vector of the previous entity $Fork$ can affect the approximated type for $Kitchen$. 



With the type context vector $\hat{\mathbf{a}}_t$ computed for each entity $e_t$, the LSTM sequentially encodes these vectors. The last hidden state of the LSTM is used as a vector representation of all selected entities types, denoted $\bm{v}_\epsilon(\hat{p})$. We then concatenate $\bm{v}_\pi(\hat{p})$ and $\bm{v}_\epsilon(\hat{p})$ together to get the final representation of our proposed path pattern $\bm{v}_\rho(\hat{p}) = [\bm{v}_\pi(\hat{p});\bm{v}_\epsilon(\hat{p})]$.

\subsection{Attention Pooling}

After representations of all paths in $\mathcal{P}_{ik}$ are obtained using the above models, we then reason over all of the resulting information, jointly, to make the prediction of whether $(e_i, r_j, e_k)$ is true. 

Prior neural network models \cite{cvsm} and \cite{chains} use a feed-forward network to condense the vector representation for each path $p_i$ to a single value $s_i$. One of the pooling methods $f_{pool}$ -- Max, Average, Top-K, or LogSumExp -- is then applied to all $s_i$ values, combining them and then passing the result through a sigmoid function $\sigma$ to make the final prediction%
\begin{align}
	P(r_j|e_i, e_k) = \sigma (f_{pool}(s_i))\:,\: \forall s_i
\end{align} %

Compressing vector representations of paths to single values, as described above, hinders the model's ability to collectively reason about the rich contextual information in the vectors.  As a result, in our approach we introduce the use of an attention mechanism for integrating information from all paths, similar to that used to compute the type context vector. We use a trainable relation-dependent vector $\bm{u}$ to represent the relation $r_j$ we are trying to predict. Following similar steps as when computing $\hat{\mathbf{a}}_t$ from $\{\bm{v}_\tau(l_{t,1}),...,\bm{v}_\tau(l_{t,C})\}$, here we compute a vector representation of all path patterns $\hat{\mathbf{p}}$ from $\{ \bm{v}_\rho(\hat{p}_1),...,\bm{v}_\rho(\hat{p}_N) \}$ conditioned on the relation-dependent vector $\bm{u}$:%
\begin{align}
	e_{i} =& f_{att, path}(\bm{v}_\rho(\hat{p}_i), \bm{u}) \\
	\alpha_{i} =& \dfrac{exp(e_{i})}{\sum_{k=1}^{N} exp(e_{k})} \\
	\hat{\mathbf{p}} =& \sum_{i=1}^{N} \alpha_{i} \bm{v}_\rho(\hat{p}_i)
\end{align} %

Since $\hat{\mathbf{p}}$ represents all the paths carrying information from both relations and entity types with correct levels of abstraction, the probability that $r_j$ exists between $e_i$ and $e_k$ can be accurately predicted using a feed-forward network $f_{pred}$ along with a sigmoid function $\sigma$: %
\begin{align}
	P(r_j|e_i, e_k) = \sigma (f_{pred}(\hat{\mathbf{p}}))
\end{align} %

\subsection{Training Procedure}

The relation encoder, entity type encoder, and attention pooling can be trained end to end as one complete model. For predicting each relation $r_j$, we train the model using true triples and false triples in the training set as positive and negative examples, denoted $\mathcal{D}_{r_j}^{Train^{+}}$ and $\mathcal{D}_{r_j}^{Train^{-}}$ respectively, requiring all triples having relation $r_j$.  Our training objective is to minimize the negative log-likelihood:%
\begin{align}
	L = &- \sum_{(e_i, r_j, e_k) \in \mathcal{D}_{r_j}^{Train^{+}}} logP(r_j|e_i, e_k) \nonumber \\
	&+ \sum_{(e_i, r_j, e_k) \in \mathcal{D}_{r_j}^{Train^{-}}} logP(r_j|e_i, e_k)
\end{align} %
We use backpropagation to update the learnable model parameters, which are the relation embedding $\bm{v}_\delta$ (dim=$50$), entity type embedding $\bm{v}_\tau$ (dim=$50$), trainable relation-dependent vector $\bm{u}$ (dim=$50$), two LSTMs in the relation encoder and entity type encoder (dim=$150$), and feedforward networks $f_{init, c}$, $f_{init, h}$, $f_{att, type}$, $f_{att, path}$, and $f_{pred}$. We used Adam \cite{adam} for optimization with default parameters (learning rate=$1e^{-3}$, $\beta_1$=$0.9$, $\beta_2$=$0.999$, $\epsilon$=$1e^{-8}$). We trained the models fully to 50 epochs. Then we used early stopping on mean average precision as regularization. 

\section{Experimental Setup}
This section describes the datasets and baselines used for evaluating our proposed method on the fact prediction task.

\subsection{Datasets}
We evaluated our method and baseline methods on two standard datasets for knowledge base completion: FB15k-237, a subset of the commonsense knowledge graph Freebase \cite{freebase}, and WN18RR, a subset of the English lexical database WordNet \cite{wordnet}. The first section of Table \ref{tab:data_stats} shows the statistics of these two datasets. 

From all true triples in each KB $\mathcal{X}$, we built up a complete dataset for experiments $\mathcal{D} = \{ (e_i, r_j, e_k, y)|e_i, e_k\in \mathcal{E} \land  r_j \in \mathcal{R} \land y \in \{0, 1\}\}$, where $y$ indicates whether a triple is true or false. This dataset contains additional false triples that are sampled from $\mathcal{X}$ using the method based on personalized page rank\footnote{\citeauthor{sfe} show that this negative sampling method has no statistically significant effect on algorithms' performance compared to sampling with PRA, another established but less efficient method used in previous works \cite{pra,deeppath}.} (released code from \cite{sfe}). Negative sampling is necessary because both standard datasets only contain true triples. As the number of negative examples has a significant effect on algorithms' performance \cite{baselines_strike_back} and evaluation, we consistently sampled 10 negative examples for each positive one. We split the dataset into 80\% training and 20\% testing. Because path-ranking based methods typically model each relation separately, the data was further divided based on relations.

\begin{table}
\centering
\begin{tabular}{lrr}  
\toprule
  & WN18RR & FB15k-237 \\
\midrule
\# Relations       & 11  &  235   \\
\# Entities            & 40,943  &   13,545   \\
\# Relation instances    &  134,720  &   254,290    \\
\midrule
\# Relations tested  & 11  &   10   \\
Avg. \# train inst/relation  & 38,039  &   16,696    \\
Avg. \# testing inst/relation  &  11,888 &   5,219    \\
\midrule
Avg. path length  & 5.3  &   3.5    \\
Max path length  & 6  &  4     \\
Avg. \# paths per instance & 88.6  &   154.8    \\
\midrule
\# Types & 8,092 & 1,029 \\
Max height of type hierarchy  & 14  &   7    \\
Avg. height of type hierarchy  & 4.6  &   6.4    \\
\bottomrule
\end{tabular}
\caption{Statistics of the datasets, the training/testing examples, the extracted paths, and the type hierarchies.}
\label{tab:data_stats}
\end{table}

\begin{table*}[t]
\centering
\begin{tabular}{lcccc}  
\toprule
& Data Types & Pooling Method & WN18RR & FB15k-237 \\
\midrule
PRA  & Relation &  N/A   & 38.85 &  34.33   \\
SFE  & Relation & N/A   &  30.75 &  36.79  \\
Path-RNN\textsuperscript{A}  & Relation &  LogSumExp    &  67.16 &  45.64   \\
Path-RNN\textsuperscript{B}  & Relation, Type & LogSumExp  & 50.82  & 51.23    \\
Path-RNN\textsuperscript{C}  & Relation, Type, Entity  & LogSumExp & 51.08  & 52.17   \\
\midrule
APR\textsuperscript{A} & Relation, Type & Max   & 73.23  &  52.40  \\
APR\textsuperscript{B} & Relation, Type & Average & 80.30  &  54.72  \\
APR\textsuperscript{C} & Relation, Type & LogSumExp & 74.51  & 54.09   \\
APR\textsuperscript{D} & Relation, Type  & Attention   &  \textbf{84.91} &  \textbf{57.35}  \\
\bottomrule
\end{tabular}
\caption{MAP\% of APR (bottom) and baseline methods (top).}
\label{tab:results}
\end{table*}

\begin{figure*}[t]
  \centering
  \begin{subfigure}[b]{0.24\textwidth}
    \includegraphics[width=\textwidth]{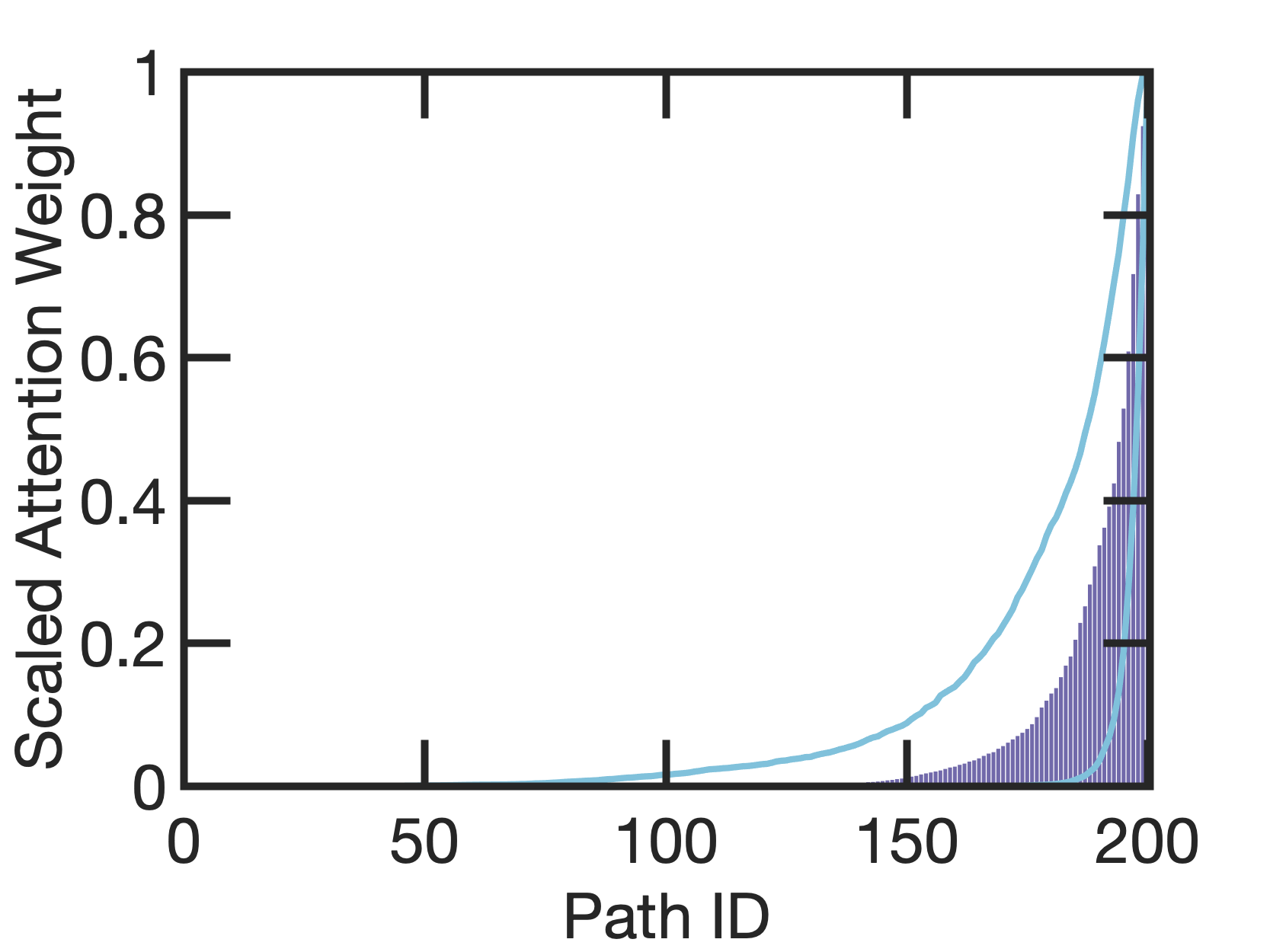}
    \caption{\scalebox{0.9}{$also\_see$}}
    \label{fig:path_weights_1}
  \end{subfigure}
  \begin{subfigure}[b]{0.24\textwidth}
    \includegraphics[width=\textwidth]{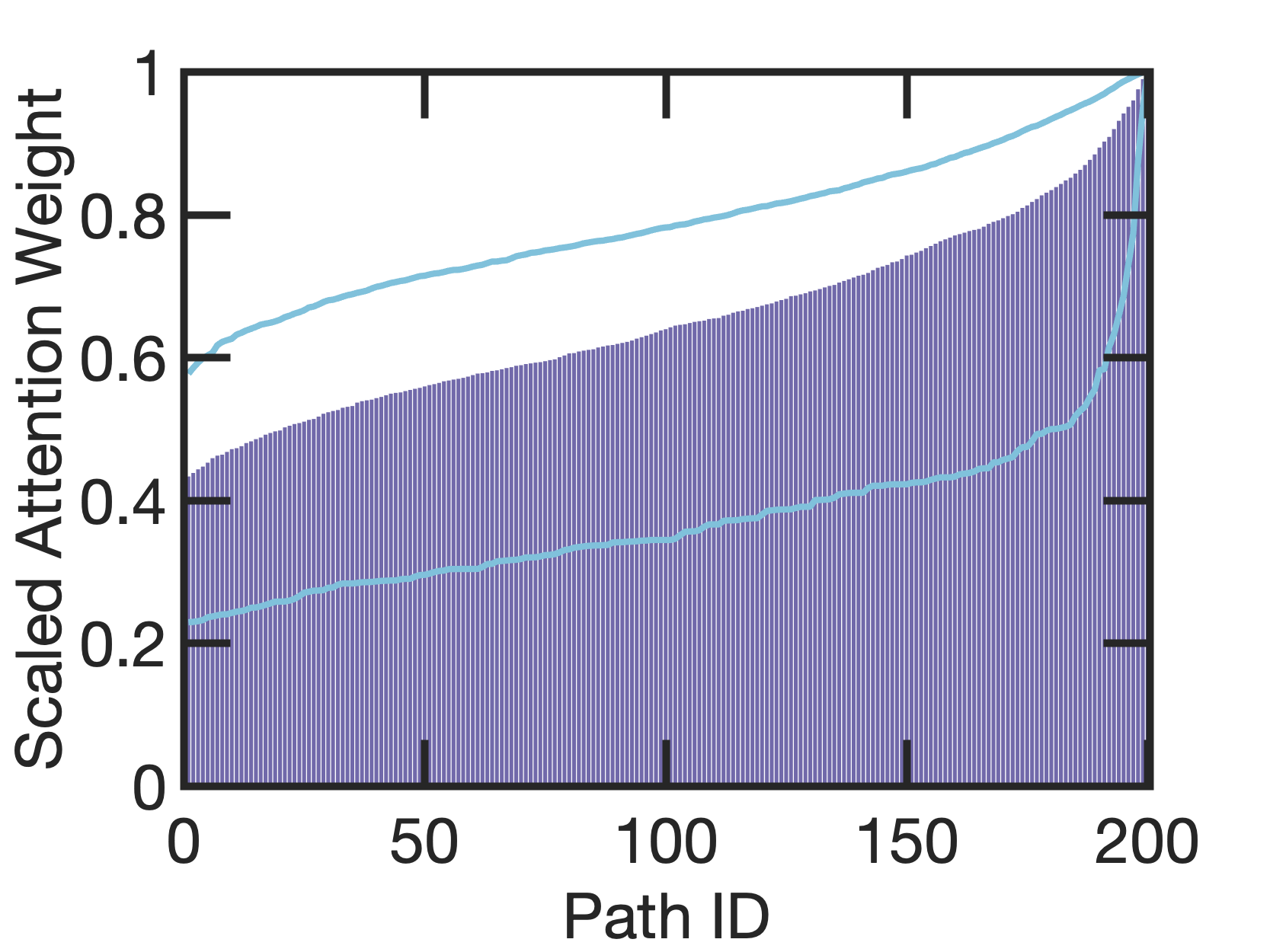}
    \caption{\scalebox{0.9}{$verb\_group$}}
    \label{fig:path_weights_2}
  \end{subfigure}
  \begin{subfigure}[b]{0.24\textwidth}
    \includegraphics[width=\textwidth]{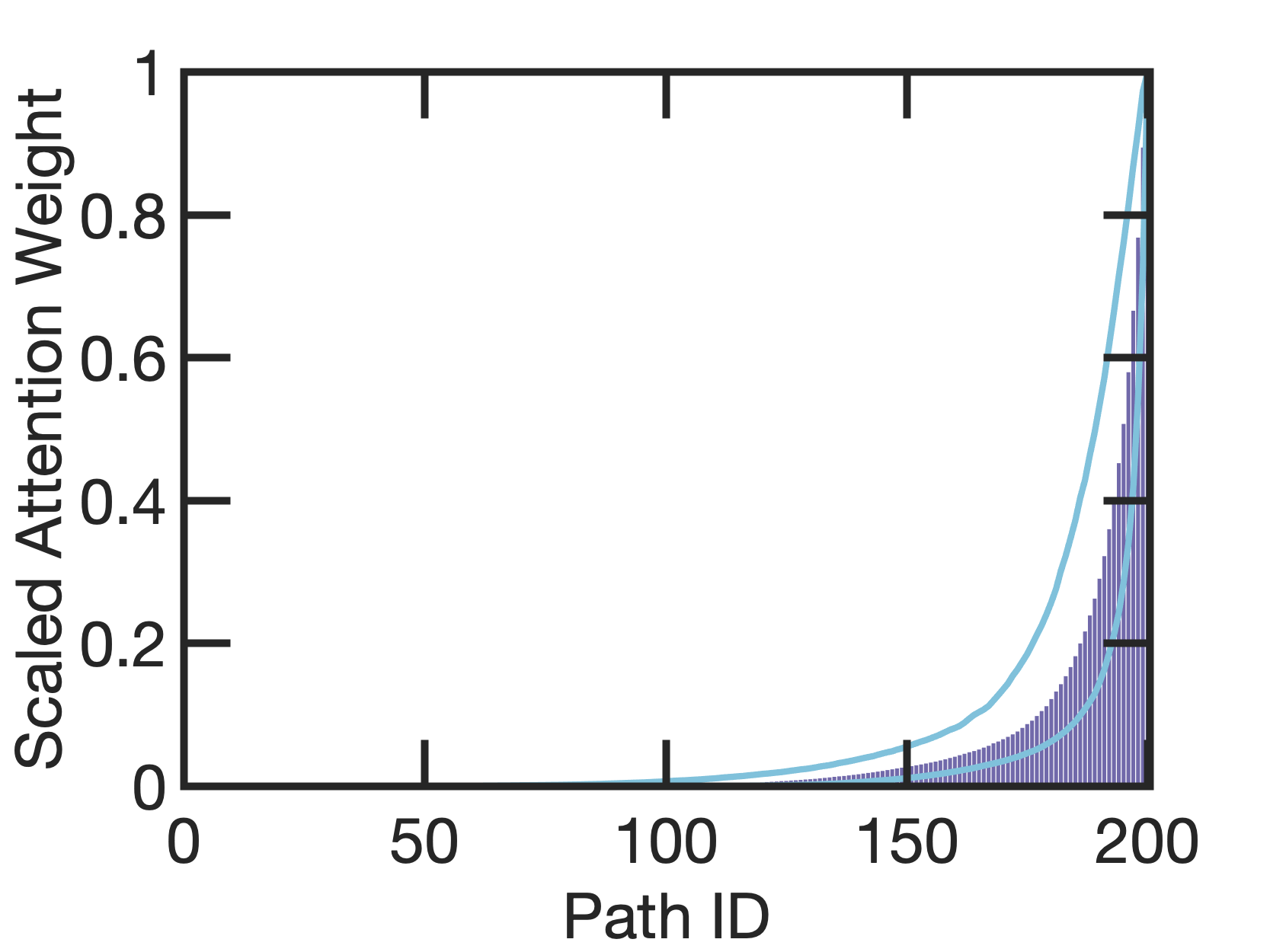}
    \caption{\scalebox{0.9}{$music/record\_label/artist$}}
    \label{fig:path_weights_3}
  \end{subfigure}
  \begin{subfigure}[b]{0.24\textwidth}
    \includegraphics[width=\textwidth]{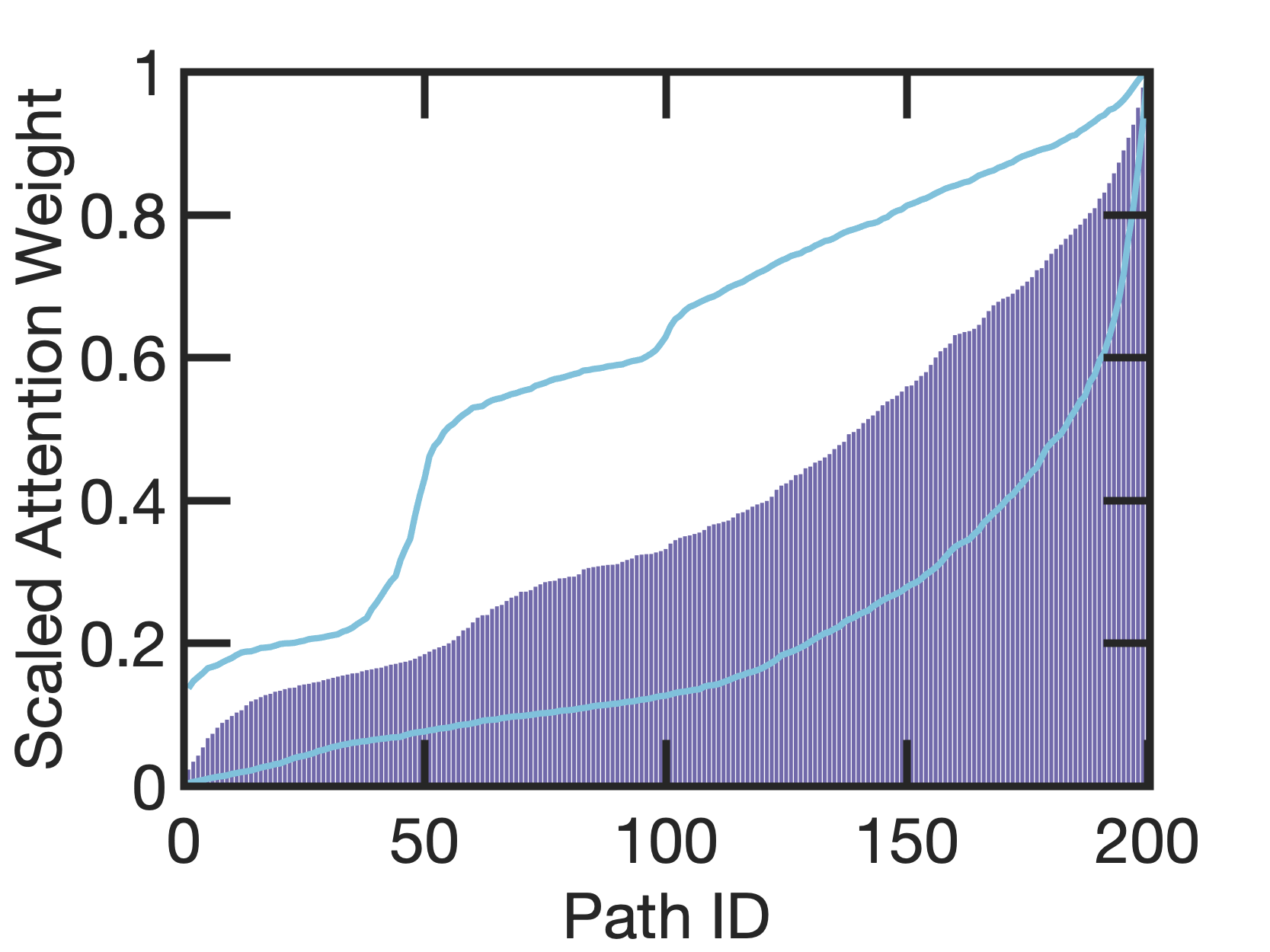}
    \caption{\scalebox{0.9}{$sports/sports\_team\_roster$}}
    \label{fig:path_weights_4}
  \end{subfigure}
  \caption{Visualization of path weight for four relations. To generate the visualization, for each relation instance, the path weights are first sorted in ascending order and normalized. Due to the varying numbers of paths between entity pairs, path weights for each relation instance are then interpolated to 200 data points. Median, 25 percentile, and 75 percentile of path weights across relation instances of each relation are plotted. }
  \label{fig:path_weights}
\end{figure*}

To extract paths between entities, we first constructed graphs from true triples in the datasets. We augmented the graphs with reverse relations following existing methods \cite{pra}. We extracted paths using bi-directional BFS. We set the maximum length of paths to 6 for WN18RR and 4 for more the densely connected FB15k-237. We randomly sampled 200 paths for each pair if there were more paths. Limiting the length of paths and sub-sampling paths are both due to computational concerns.

To create type hierarchies, we extracted inherited hypernyms available in WordNet \cite{wordnet} for WN18RR entities and used type data released in \cite{fb_types} for FB15K-237 entities. The number of type levels was automatically determined by the number of inherited hypernyms for each entity in WN18RR and the number of types for each entity in FB15K-237. We further ordered Freebase types based on their frequency of occurrence because types for Freebase entities are not strictly hierarchical. We then followed \cite{chains} to select up to 7 most frequently occurring types for each entity. As more specific types apply to fewer entities and therefore appear in fewer entities’ types, ordering types by frequencies allowed us to construct hierarchies of types with specific types generally in the bottom levels and abstract types generally in the top levels. Finally, for WN18RR, we mapped types to their vector representations using a pre-trained Google News word2vec model. Because we did not find a suitable pre-trained embedding for types of FB15K-237 entities, we trained an embedding with the whole model end-to-end. 

\subsection{Baselines}
We compared the performance of APR to the following  methods\footnote{The first two methods are tested with code released by \citeauthor{sfe}, and the other three with code by \citeauthor{chains}. All baselines have been tuned on the validation data and the result from the best hyperparameters is reported for each model.}\footnote{The first two baselines implement path extraction themselves, other methods and our models use the paths we extracted.}: \textit{\textbf{PRA}} from \cite{pra}, \textit{\textbf{SFE}} from \cite{sfe}, \textit{\textbf{Path-RNN\textsuperscript{A}}} from \cite{cvsm}, and the two models \textit{\textbf{Path-RNN\textsuperscript{B}}} and \textit{\textbf{Path-RNN\textsuperscript{C}}} from \cite{chains}. \textit{\textbf{Path-RNN\textsuperscript{B}}} and \textit{\textbf{Path-RNN\textsuperscript{C}}} are different in the data types they use, as shown in Table \ref{tab:results}.


    





\section{Results}
In this section, we report results comparing performance to the above prior methods, as well as independent validation of attention-based pooling.  We additionally present insights into the APR model and its modeling of abstraction.

\subsection{Comparison to Existing Methods}
We compared the performance of APR, paired with various pooling methods, against the baselines; Table \ref{tab:results} summarizes the results. All \textit{\textbf{APR}} variants outperformed the prior state of the art on both datasets. Directly comparing all methods that utilize LogSumExp pooling (\textit{\textbf{APR\textsuperscript{C}}} and all three \textit{\textbf{Path-RNN}} models), our results show statistically significant\footnote{As determined by a paired t-test, with each relation treated as paired data. ($p<0.05$)} improvement of \textit{\textbf{APR\textsuperscript{C}}}.  This result indicates that adding types, with the right balance between being generalizable and discriminative, helps create path patterns that allow for more accurate prediction. Our best model \textit{\textbf{APR\textsuperscript{D}}}, which further leverages attention pooling to reason about the rich contextual information in paths, is able to improve state-of-the-art performance by $26\%$ on WN18RR and $10\%$ on FB15k-237 with~($p<0.005$).

One surprising result is that \textit{\textbf{Path-RNN\textsuperscript{B}}} and \textit{\textbf{Path-RNN\textsuperscript{C}}}, even using entity and type information, still perform worse than \textit{\textbf{Path-RNN\textsuperscript{A}}} on WN18RR. We suspect that the extremely large number of entities and types in WN18RR, and the simple feature aggregation method used by these models, cause learning to not generalize even for highly adaptable neural network models. The use of abstraction helps our model generalize information from individual entities and types, and achieve more robust prediction.
%
%
\subsection{Pooling Methods}
We compared our proposed attention pooling to three existing pooling methods. As shown in Table \ref{tab:results}, \textit{\textbf{APR\textsuperscript{D}}} with attention pooling performs the best on both datasets. The superior performance is likely due to the pooling method's ability to collectively reason about the rich contextual information in paths. The other three methods lose information when they compress path representations to single values. 

To gain insight about the behavior of attention pooling, we visualized the path weights $\alpha_{i}$'s computed by the attention module. Figure \ref{fig:path_weights} shows visualizations of four representative relations from the two datasets. Figure \ref{fig:path_weights_1} and Figure \ref{fig:path_weights_3} show that for some relations, attention pooling focuses on small numbers of highly predictive paths. In other cases, as in Figures \ref{fig:path_weights_2} and \ref{fig:path_weights_4}, attention pooling incorporates data across a broad collection of paths. This ability to dynamically adjust attention based on context highlights the adaptability of attention pooling. 

\begin{table}[t]
\centering
\begin{tabular}{lcc}  
\toprule
& WN18RR & FB15k-237 \\
\midrule
Abstract   & 81.17 &  49.71  \\
Specific      & 82.66  &  54.55 \\
Attention (APR\textsuperscript{D}) & \textbf{84.91} & \textbf{57.35} \\
\bottomrule
\end{tabular}
\caption{MAP\% of three models using different levels of abstraction for entities.}
\label{tab:attention_effect}
\end{table}

\subsection{Correct Level of Abstraction}

\begin{table*}[h]
\centering
\resizebox{\textwidth}{!}{%
\begin{tabular}{lccccccc}  
\toprule
\multicolumn{8}{l}{Query Relation: has\_profession}\\
\multicolumn{8}{l}{Interpretation: Actors graduated from the same school are likely to share the same profession.}\\
&&&&&&&\\
True:&$\overset{\scalebox{1}{[Actor]}}{\text{John
Rhys-Davis}}$&$\xleftarrow{\scalebox{1}{has\_graduates}}$&$\overset{\scalebox{1}{[Thing]}}{\text{RADA}}$&$\xrightarrow{\scalebox{1}{has\_graduates}}$&$\overset{\scalebox{1}{\textbf{[Actor]}}}{\text{Micheal Gambon}}$&$\xrightarrow{\scalebox{1}{has\_profession}}$&$\overset{\scalebox{1}{[Thing]}}{\text{Actor}}$\\
False:&$\overset{\scalebox{1}{[Actor]}}{\text{Peter Ustinov}}$&$\xleftarrow{\scalebox{1}{has\_graduates}}$& $\overset{\scalebox{1}{[Thing]}}{\text{Westminster School}}$&$\xrightarrow{\scalebox{1}{has\_graduates}}$&$\overset{\scalebox{1}{\textbf{[Author]}}}{\text{Christopher Wren}}$&$\xrightarrow{\scalebox{1}{has\_profession}}$&$\overset{\scalebox{1}{[Thing]}}{\text{Scientist}}$\\
\midrule
\multicolumn{8}{l}{Query Relation: has\_genre}\\
\multicolumn{8}{l}{Interpretation: Films focusing on the same subject are likely to share the same genre. However, films depicting the same profession can have different genres.}\\
&&&&&&&\\
True:&$\overset{\scalebox{1}{[Topic]}}{\text{Life Is Beautiful}}$&$\xleftarrow{\scalebox{1}{has\_film}}$&$\overset{\scalebox{1}{\textbf{[Subject]}}}{\text{World War 2}}$&$\xrightarrow{\scalebox{1}{has\_film}}$&$\overset{\scalebox{1}{[Topic]}}{\text{Valkyrie}}$&$\xrightarrow{\scalebox{1}{has\_genre}}$&$\overset{\scalebox{1}{[Genre]}}{\text{War Film}}$\\
False:&$\overset{\scalebox{1}{[Topic]}}{\text{Air Force One}}$&$\xleftarrow{\scalebox{1}{has\_film}}$&$\overset{\scalebox{1}{\textbf{[Profession]}}}{\text{Aviation}}$&$\xrightarrow{\scalebox{1}{has\_film}}$&$\overset{\scalebox{1}{[Topic]}}{\text{Team America}}$&$\xrightarrow{\scalebox{1}{has\_genre}}$&$\overset{\scalebox{1}{[Genre]}}{\text{Political Satire}}$\\
\midrule
\multicolumn{8}{l}{Query Relation: nominee\_work}\\
\multicolumn{8}{l}{Interpretation: The best actors are likely to perform great in all movies. The same company may not always produce award winning works.}\\
&&&&&&&\\
True:&$\overset{\scalebox{1}{\textbf{[Thing]}}}{\text{Academy Actor}}$&$\xrightarrow{\scalebox{1}{nominee\_work}}$&$\overset{\scalebox{1}{\textbf{[Award Topic]}}}{\text{Network}}$&$\xleftarrow{\scalebox{1}{nominated\_for}}$&$\overset{\scalebox{1}{\textbf{[Actor]}}}{\text{Sidney Lumet}}$&$\xrightarrow{\scalebox{1}{nominated\_for}}$&$\overset{\scalebox{1}{\textbf{[Award Item]}}}{\text{Orient Exp}}$\\
False:&$\overset{\scalebox{1}{\textbf{[Award Category]}}}{\text{Emmy Comedy}}$&$\xrightarrow{\scalebox{1}{nominee\_work}}$&$\overset{\scalebox{1}{\textbf{[Tv Program]}}}{\text{Entourage}}$&$\xleftarrow{\scalebox{1}{nominated\_for}}$&$\overset{\scalebox{1}{\textbf{[Employer]}}}{\text{HBO}}$&$\xrightarrow{\scalebox{1}{nominated\_for}}$&$\overset{\scalebox{1}{\textbf{[Tv Program]}}}{\text{Gia}}$\\
\bottomrule
\end{tabular}}
\caption{Correctly predicted examples by the proposed model on test data of FB15k-237. Despite the fact that each pair of true and false examples share the same sequence of relations, the uses of entity types at the right level of abstraction (shown in bold) help distinguish between them. Abbreviated entity and relation names are shown for clarity.}
\label{tab:qualitative_result}
\end{table*}

We further investigated whether our model learns the proposed path patterns that balance between discrimination and generalization. For comparison, we modified the best model \textit{\textbf{APR\textsuperscript{D}}} by fixing the selection of types to either the most specific level or the most abstract level. Table \ref{tab:attention_effect} shows the effect of different levels of abstraction on performance. Using a fixed level of abstraction leads to worse performance compared to using attention. This result not only confirms that balancing between generalization and discrimination makes prediction more accurate, but also that our proposed model achieves this balance by learning the correct levels of abstraction for entities.

\begin{figure}[t]
  \centering
  \begin{subfigure}[b]{0.23\textwidth}
    \includegraphics[width=\textwidth]{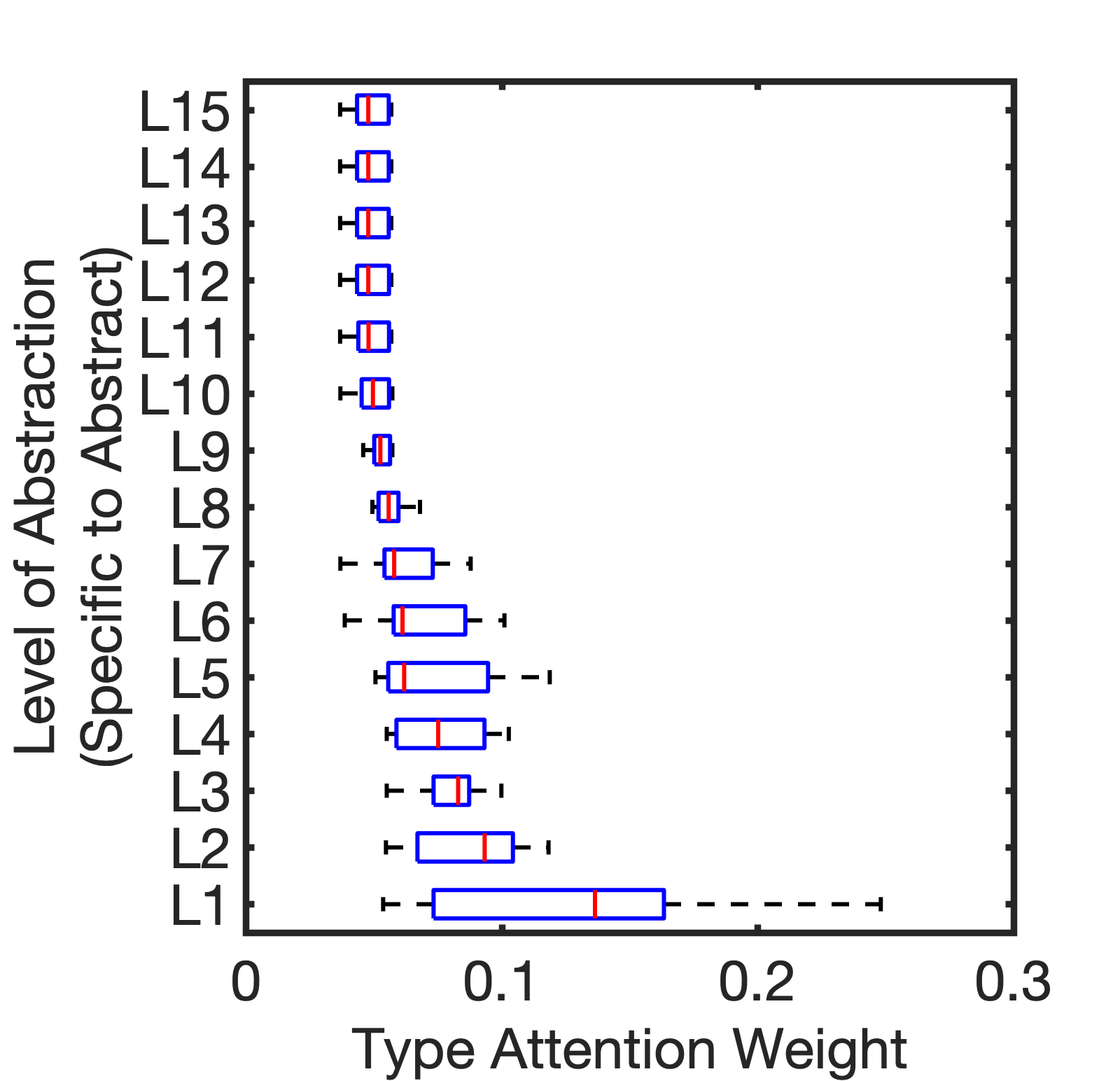}
    \caption{Average type weights of WN18RR}
    \label{fig:type_weights_wn}
  \end{subfigure}
  \begin{subfigure}[b]{0.23\textwidth}
    \includegraphics[width=\textwidth]{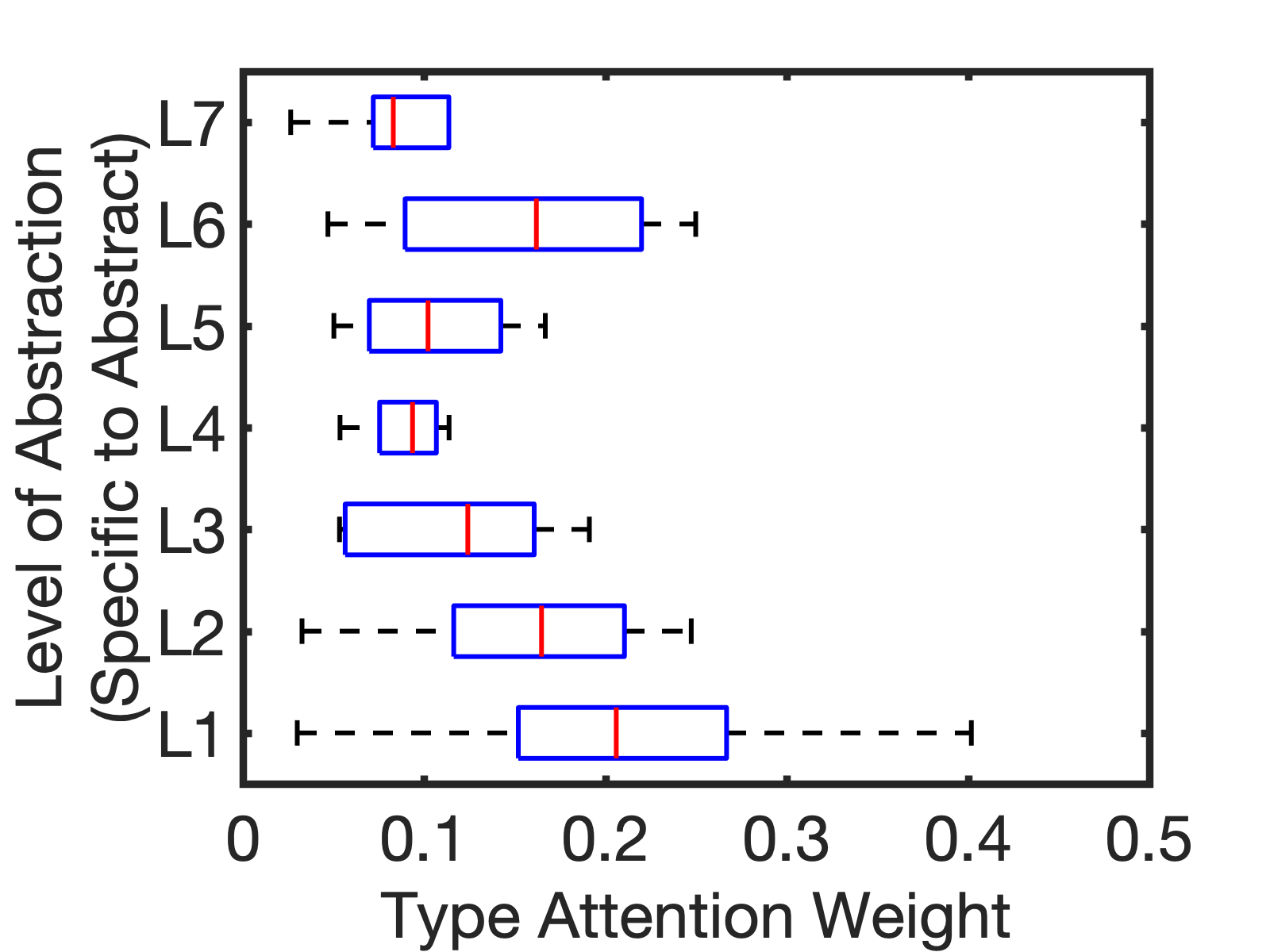}
    \caption{Average type weights of FB15k-237}
    \label{fig:type_weights_fb}
  \end{subfigure}
  \caption{Box plot of average attention weights at each level. For each relation, the average attention weight of each level is computed by averaging over the weights predicted by the model at this level for all entities in all paths.}
  \label{fig:type_weights}
\end{figure}

We also examined the distribution of attention weight at different levels of type hierarchies. Figure \ref{fig:type_weights} shows that the model leveraged all levels for representing various entities. More specifically, level 1, 2, and 3 are most commonly emphasized for entities in WN18RR; level 1,2, and 6 are most often used for entities in FB15k-237. The emphasis on lower levels is expected since using the most specific level is more beneficial than using the most abstract level with evidence in Table \ref{tab:attention_effect}. However, the diversity of levels used by the model proves that not all entities should use the most specific level: different entities require different levels of abstraction.

Finally, we visualized examples of paths from prediction along with the top weighted types the model selected for entities. In Table \ref{tab:qualitative_result}, we are able to see examples that correspond strongly with human intuition. This qualitative result again verifies that the model learns a new class of rules (path patterns) that is more specific yet still generalizable. 

\begin{table}[t]
\centering
\begin{tabular}{lcc}  
\toprule
& WN18RR & FB15k-237 \\
\midrule
TuckER   & 38.16 &  40.12  \\
ComplEx-N3     & 40.89  &  40.35 \\
APR\textsuperscript{D}      & \textbf{84.91} & \textbf{57.35} \\
\bottomrule
\end{tabular}
\caption{MAP\% comparison to two SOTA embedding methods.}
\label{tab:comparison_with_embedding_methods}
\end{table}

\subsection{Comparison with Knowledge Graph Embedding Methods}
As a final point of comparison, we validated our best model \textit{\textbf{APR\textsuperscript{D}}} against two state-of-the-art KGE methods\footnote{We followed \cite{deeppath} to evaluate KGE methods on the fact prediction task. Test triples with the same relation are ranked.}: \textit{\textbf{TuckER}} \cite{tucker} and \textbf{\textit{ComplEx-N3}} \cite{complex}. As shown in Table \ref{tab:comparison_with_embedding_methods}, our model performs significantly better on both datasets. In fact, all neural-based path ranking methods (\textit{\textbf{APR}} and \textbf{\textit{Path-RNN}}) outperform the KGE methods. One possible reason that path ranking methods outperform KGE is that KGE methods require both the source $e_i$ and target entities $e_k$ in the missing triple $(e_i, r_j, e_k)$ to be known, thus failing to perform well when test entities are not present in the training set.  However, only $17.6039\%$ of FB15k-237's test data consists of new entities, and only $0.9412\%$ in WN18RR.  As a result, new entities do not alone account for the difference in performance.  This comparison additionally affirms the improved state-of-the-art performance in fact prediction achieved by our method.

\section{Conclusion}
This work addresses the problem of knowledge base completion.  We introduced Attentive Path Ranking, a novel class of generalizable path patterns leveraging type hierarchies of entities, and developed an attention-based RNN model to discover the new path patterns from data. Our approach results in statistically significant improvement over state-of-the-art path-ranking based methods and knowledge graph embedding methods on two benchmark datasets WN18RR and FB15k-237. Quantitative and qualitative analyses of the discovered path patterns provided insights into how APR achieves a balance between generalization and discrimination.

\section{ Acknowledgments}
This work is supported in part by NSF IIS 1564080, NSF GRFP DGE-1650044, and ONR N000141612835.

\bibliographystyle{aaai}
\bibliography{AAAI-LiuW.2859}

\end{document}